\documentclass[runningheads]{llncs}
\usepackage[T1]{fontenc}
\usepackage{graphicx}
\begin{document}
\title{Characterizing the Features of Mitotic Figures Using a Conditional Diffusion Probabilistic Model}

\titlerunning{Characterizing the Features of Mitotic Figures Using a Conditional DPM}
%
\author{Cagla Deniz Bahadir 
\inst{1}\orcidID{0000-0003-2247-9975} 
\and Benjamin Liechty 
\inst{2} \orcidID{0000-0001-6485-5604} 
\and David J. Pisapia 
\inst{2}  \orcidID{0000-0003-1528-8425} 
\and Mert R. Sabuncu 
\inst{3,4} \orcidID{0000-0002-7068-719X} } 
\authorrunning{C.D. Bahadir et al.}
%
\institute{Cornell University and Cornell Tech, Biomedical Engineering, New York, NY, USA \and Weill Cornell Medicine, Pathology and Laboratory Medicine, New York, NY, USA \and Weill Cornell Medicine, Radiology, New York, NY USA \and Cornell University and Cornell Tech, Electrical and Computer Engineering, New York, NY, USA}
\maketitle              
\begin{abstract}
Mitotic figure detection in histology images is 
a hard-to-define, yet clinically significant task, where labels are generated with pathologist interpretations and where there is no ``gold-standard'' independent ground-truth. 
However, it is well-established that these interpretation based labels are often unreliable, in part, due to differences in expertise levels and human subjectivity. In this paper, our goal is to shed light on the inherent uncertainty of mitosis labels and characterize the mitotic figure classification task in a human interpretable manner. We train a probabilistic diffusion model to synthesize patches of cell nuclei for a given mitosis label condition. Using this model, we can then generate a sequence of synthetic images that correspond to the same nucleus transitioning into the mitotic state. This allows us to identify different image features associated with mitosis, such as cytoplasm granularity, nuclear density, nuclear irregularity and high contrast between the nucleus and the cell body. Our approach offers a new tool for pathologists to interpret and communicate the features driving the decision to recognize a mitotic figure.

\end{abstract}

\section{Introduction}

Mitotic figure (MF) count is an important diagnostic parameter in grading cancer types including meningiomas \cite{ganz2021automatic}
, breast cancer \cite{sohail2021multi,veta2013detecting}, uterine cancer \cite{zehra2022novel}, based on criteria set by World Health Organization (WHO) 
\cite{ganz2021automatic,sohail2021multi,zehra2022novel}. Due to the size of the whole slide images (WSIs), scarcity of MFs, and that presence of cells that mimic the morphological features of MFs, the detection process by pathologists is time consuming, subjective, and prone to error. This has led to the development of algorithmic methods that try to automate this process~\cite{saha2018efficient,wu2017ff,sebai2020partmitosis,bertram2019large,aubreville2020deep}.

Earlier studies have used classical machine learning models like SVMs or random forests \cite{albayrak2016mitosis,beevi2019automatic} to automate MF detection.
Some of these works relied on morphological, textural and intensity-based features~\cite{saha2018efficient,sigirci2022detection}. 
More recently, modern deep learning architectures for object detection, such as RetinaNet and Mask-RCNN have been employed for candidate selection, which is in turn followed by a ResNet or DenseNet-style classification model for final classification \cite{wilm2022domain,fick2022domain}.
Publicly available datasets and challenges such as Canine Cutaneous Mast Cell Tumor (CCMCT) \cite{bertram2019large} and MIDOG 2022 \cite{aubreville2023mitosis}, have catalyzed research in this area.

An important aspect of MF detection is that there is no independent ground-truth, other than human-generated labels. However, high inter-observer variability of human annotations has been documented~\cite{malon2012mitotic}. In a related paper~\cite{bertram2020pathologist}, the examples in the publicly available TUPAC16 dataset were re-labeled by two pathologists with the aid of an algorithm, which revealed that the updated labels can markedly change the resultant F1-scores. In another study, pathologists found the lack of being able to change the z axis focus, a limiting factor for determining MFs in digitized images~\cite{wei2019agreement}. Despite high inter-observer variability and inherent difficulty of recognizing MFs, all prior algorithmic work in this area relies on discrete labels, e.g. obtained via a consensus \cite{albarqouni2016aggnet,aubreville2023mitosis,bertram2019large}. 
One major area that hasn't been studied is the morphological features of the cell images that contribute to the uncertainty in labeling MF. 

Diffusion probabilistic models (DPMs) have recently been used to generate realistic images with or without conditioning on a class \cite{ho2020denoising}, including in biomedical applications ~\cite{moghadam2023morphology,sanchez2022healthy}. In this paper, we trained a conditional DPM to synthesize cell nucleus images corresponding to a given mitosis score (which can be probabilistic). The synthetic images were in turn input to a pre-trained MF classification model to validate that the generated images corresponded to the conditioned mitosis score. The prevailing MF classification literature considers MF labels as binary. Embracing a probabilistic approach with the help of DPMs offers the opportunity to characterize the MF features, thus improves the interpretability of the MF classification process. Our DPM allows us to generate synthetic cell images that illustrate the transition from a non-mitotic figure to a definite mitosis. We also present a novel approach to transform a real non-mitotic cell into a mitotic version of itself, using the DPM. Overall, these analyses allow us to reveal and specify the image features that drive the interpretation of expert annotators and explain the sources of uncertainty in the labels.

\section{Methodology}

\subsection{Datasets}

\textbf{Canine Cutaneous Mast Cell Tumor:}
CCMCT is a publicly available dataset that comprises 32 whole-slide images (WSI)~\cite{bertram2019large}. We used the ODAEL (Object Detection Augmented Expert Labeled) variant which had the highest number of annotations. The dataset has a total of 262,481 annotated nuclei (based on the consensus of two pathologists). 44,880 of them are mitotic figures and the remainder are negative examples. We used the same training and test split as the original paper: 21 slides for training and 11 slides for test. \\
\textbf{Meningioma:}
We created a meningioma MF dataset using 7 WSIs from a public dataset (The Digital Brain Tumour Atlas \cite{roetzer2022digital,thomas2022digital2}) and 5 WSIs from our own hospital. The WSIs were non-exhaustively annotated for mitotic figures by two pathologists. The pathologists were allowed to give a score of 0, 0.5 and 1 to each candidate cell, 0.5 representing possible MFs where the annotator was not fully certain. There are a total of 6945 annotations, with 4186 negatives, 1439 definite MFs and 1320 annotations with scores varying between 0 and 1. 

\begin{figure}[t]
\includegraphics[width=\textwidth]{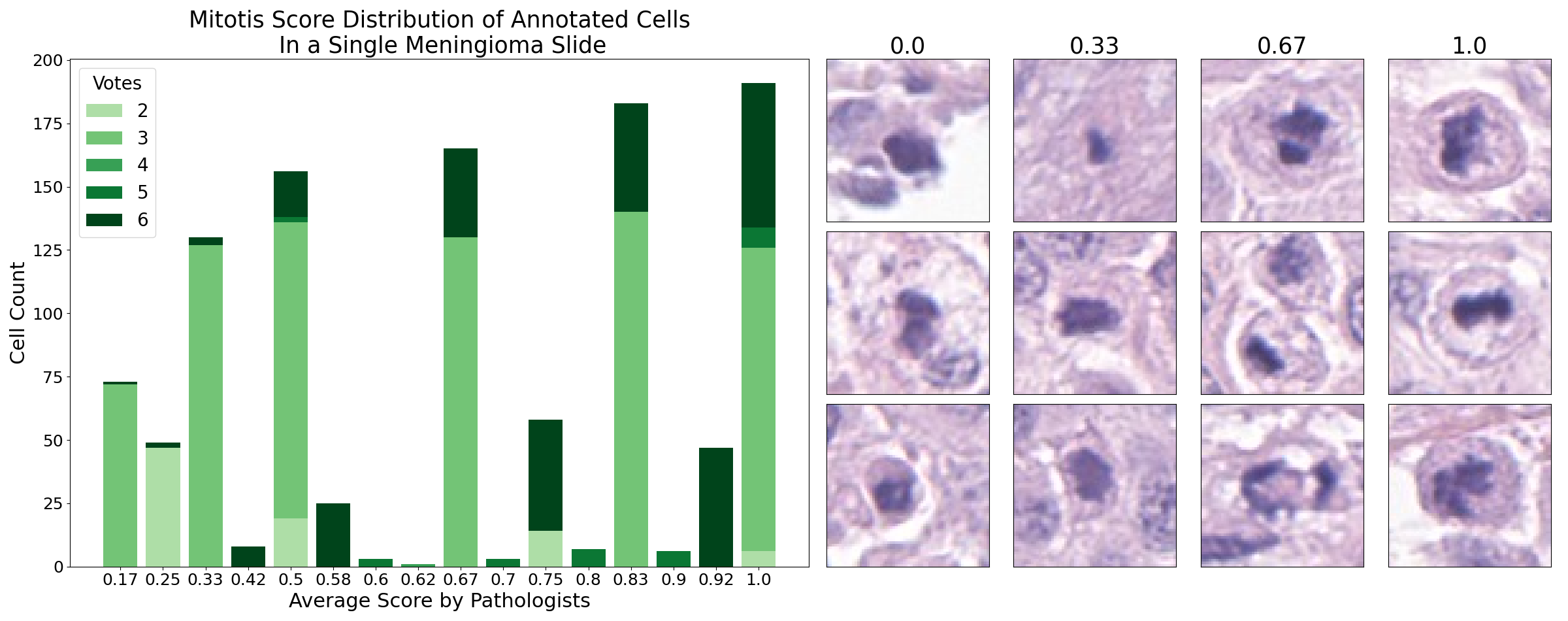}
\caption{Label distribution from a meningioma slide (left) and real cell images from a meningioma slide labeled with pathologists’ scoring (right).} \label{fig1}
\end{figure}

For 11 slides in the dataset each cell was seen by one of our two pathologists. 
For one slide in the dataset a total of 2095 cells were annotated by two pathologists using an in-house software that showed the cells to the annotators repetitively in a shuffled manner. The 2095 cells were independently annotated by both pathologists, up to three times each, yielding 6138 annotation instances. Fig \ref{fig1} (left) shows the distribution of scores given to cells that are possibly MFs (i.e. have a non-zero score). The bar plot has also been color coded to show how many votes have been gathered for each score. 
This figure clearly demonstrates the probabilistic nature of the ground-truth label. Fig.~\ref{fig1}-right shows 12 cell examples with 4 different scores. Each of these cells has been annotated at least 3 times. The uncertainty of the scoring can be explained by visual inspection. 
In column 0, the pathologists are in consensus that these examples are not mitotic. In the top two examples, cytoplasmic membranes are indistinct or absent and the basophilic material that might prompt consideration of a mitotic figure does not have the shape or cellular context of condensed chromatin that would be encountered in a MF. The bottom example is a cell with a hyperchromatic nucleus that is not undergoing mitosis. In the examples with a 0.33 score, we observe that there is some density in the chromatin which can confound the interpreter. Yet, the lack of cellular borders make it challenging to interpret the basophilic material, and the putative chromatin appears smoother and lighter compared to the more confident examples. The second row in 0.67 is likely an example of telophase, and the uncertainty is attributable to  the lighter and blurry appearance of the upper daughter cell. Also the potential anaphase in the third row of 0.67 shows an indistinct to absent cytoplasmic membrane, introducing uncertainty for a definitive mitotic figure call. The cell examples in column 1.0 depict strong features such as high chromatin density and irregularity in the nucleus and are good examples of metaphase due to the elongated appearance.

\subsection{Training}

We trained a ResNet34 model on the CCMCT dataset training set (21 WSIs) to discriminate mitotic and non-mitotic cells. The model was initialized with ImageNet pretrained weights. We used the Adam optimizer with a 1e-05 learning rate, batch size of 128, and binary cross entropy loss. The mitotic figures were sampled randomly in each mini batch with an equal number of randomly sampled negative examples. 64x64 RGB tiles were created around each cell nucleus and re-sampled to 256x256 to match Resnet34's input requirement. Each WSI was used in training and validation, with a 75\% division in the y axis. 
The model was trained until validation loss converged. We randomly initialized with 3 different seeds during training and ensembled the three converged models by averaging to obtain final scores. 
This ResNet34 ensemble reached 0.90 accuracy and 0.81 F1-score in test data, similar to the numbers reported in the CCMCT paper with ResNet18 \cite{bertram2019large}.

For the conditional DPM, we used an open-source PyTorch implementation\footnote{https://github.com/lucidrains/denoising-diffusion-pytorch}. The input embedding layer was changed into a fully connected layer that accepts a real-valued (probabilistic) input. 
Two DPMs were trained. One model was trained on the CCMCT dataset with the 11 test slides that were not seen by the classification model during training. Importantly, the CCMCT includes binary (i.e., deterministic) labels.
The second DPM was trained on the meningioma dataset that comprised 12 slides with probabilistic labels. The models were trained on 64x64 images, with Adam optimizer, learning rate of 1e-04 and a batch size of 128. 
The trained model weights, code, and our expert annotations for The Brain Tumor Atlas Dataset is made publicly available \footnote{https://github.com/cagladbahadir/dpm-for-mitotic-figures}.

\begin{figure}[t]
\centering
\includegraphics[width=0.7\textwidth]{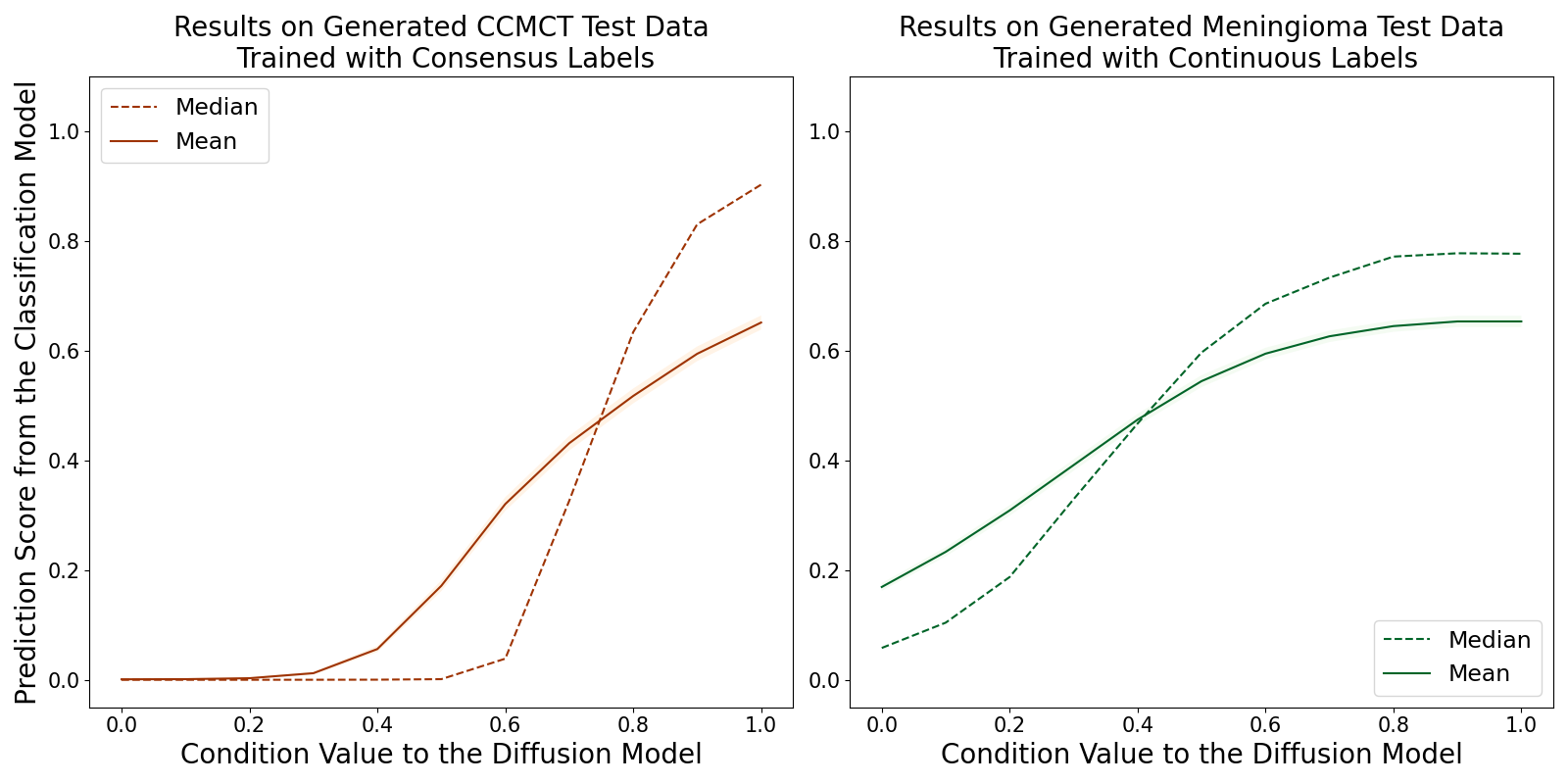}
\caption{Generated images from the diffusion model is tested on the ResNet34 classification model. The plot on the left shows the ResNet34 scores on the CCMCT generated test data and the plot on the right shows the scores for the synthetic meninigoma data. 
The shaded regions correspond to the standard error of the distribution of classification scores for every condition value. } \label{fig2}
\end{figure}

\subsection{Inference}

During inference time, for a given random noise seed input, the DPM was run with a range of scores between 0 and 1, at 0.1 increments, as the conditional input value. This way, we generated a sequence of synthetic cell images that corresponds to a random generated cell nucleus transitioning into the mitosis stage.

We were also interested in using the DPMs to visualize the transformation of a \textit{real} non-mitotic cell nucleus into a MF.
To achieve this, we ran the DPM in forward diffusion mode, starting from a real non-mitotic cell image and iteratively adding noise.
At intermediate time-points, we would then stop and invert the process to run the DPM to denoise the image - this time, conditioning on an MF score of 1 (i.e., definite mitosis).
This allowed us to generate a sequence of images, each corresponding to different stopping time-points.
Note that earlier stopping time-points yield images that look very similar to the original input non-mitotic image.
However, beyond a certain time-point threshold, the synthesized image looks like a mitotic version of the input. 
We can interpret this threshold as a measure of how much an input image resembles a mitotic figure.

\section{Results}

\subsection{Generated Cells with Probabilistic Labels}

1000 sets of cells were generated with the two DPMs. Each set is a series of 11 synthetic images generated with the same random seed and condition values between 0 and 1, with 0.1 increments. 
The generated images were then input to the ResNet34 classification model. Fig \ref{fig2} visualizes the average ResNet34 model scores for synthetic images generated for different condition values. 

The CCMCT results presented on the left shows a steep increase in median classification scores given by the prediction model, starting from the conditional value of 0.6. The meningioma model which was trained on probabilistic labels, shows a more gradual trend of increase in the classification scores.
This difference is likely due to fact that the CCMCT DPM was trained with binary labels and thus was not exposed to probabilistic conditional values, whereas the meningioma DPM was trained with probabilistic scores. 
The utilization of continuous labels obtained from pathologist votes during the training of DPMs enhances the comprehension of underlying factors that contribute to labeling uncertainty.

\begin{figure} [t]
\centering
\includegraphics[width=\textwidth]{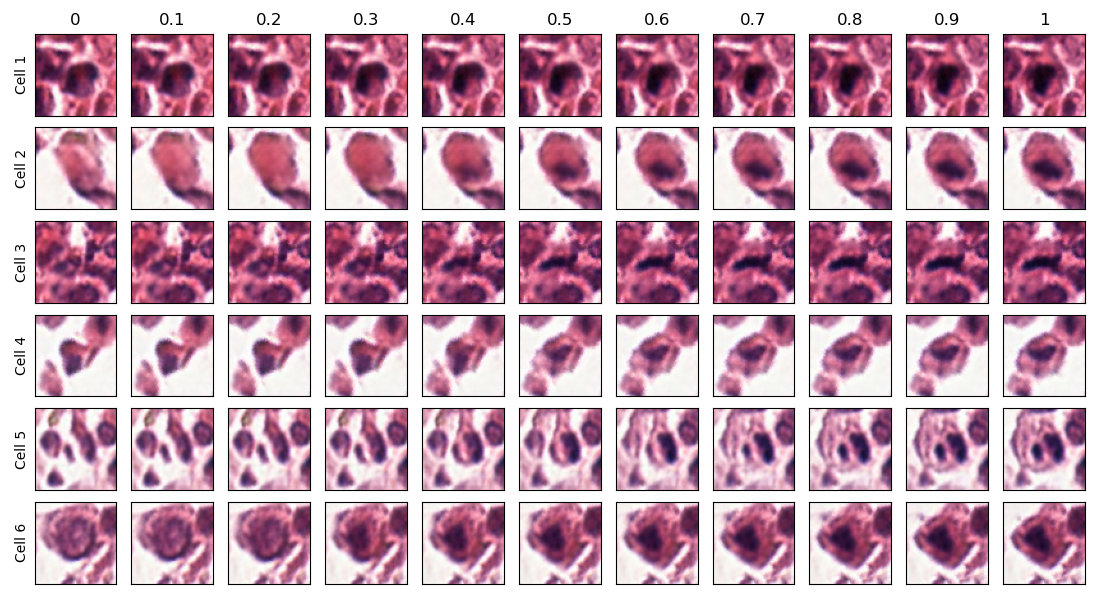}
\caption{Synthetic cell sets generated with the CCMT DPM. Rows represent cell sets generated from the same random noise, transitioning from non-mitotic (left) to definite mitotic figures (right). Condition values are listed above.} \label{fig3}
\end{figure}

\subsection{Selecting Good Examples}

In order to select good synthetic images to interpret, we passed each set to the classification model. The identification of "good synthetic" images is contingent upon task definition and threshold selection. In our study, we adopted a selection process focused on visually evaluating smooth transitions between synthetic images. If a synthetic image set started with predicted mitosis score of less than 0.1, reached a final score greater than 0.9 and the sequence of scores were relatively smooth (e.g., the change in scores was less than 0.30 between each increment), we included it in our visual analysis presented below.

\subsection{CCMCT Visual Results}

Fig \ref{fig3} shows the selected examples from the diffusion model trained on the CCMCT dataset. Each row represents one set that was generated with the same random seed. This visual depiction shows that there are a wide variety of morphological changes that correlate with transitioning from a non-mitotic to a mitotic cell. In cells 1 and 5 there are several dark areas at lower probabilities in the first image. The model merges several of the spots to one, bigger and darker chromosomal aggregate in cell 1 and two dense chromosomal aggregates in cell 5, mimicking telophase. The surrounding dark spots are faded away and the cell membrane becomes more defined. In cells 2 and 6, condensed chromatin is generated from scratch as the conditional value is increased. In cell 3, the faint elongated structure in the center of the image gradually merges into a denser, more defined, irregular and elongated chromosomal aggregate, mimicking metaphase. In cell 4, the cell body size increases around the nucleus while gradually creating more contrast by lowering the darkness of the cytoplasm, sharpening the edges and increasing the density of the chromatin.

\begin{figure}
\centering
\includegraphics[width=\textwidth]{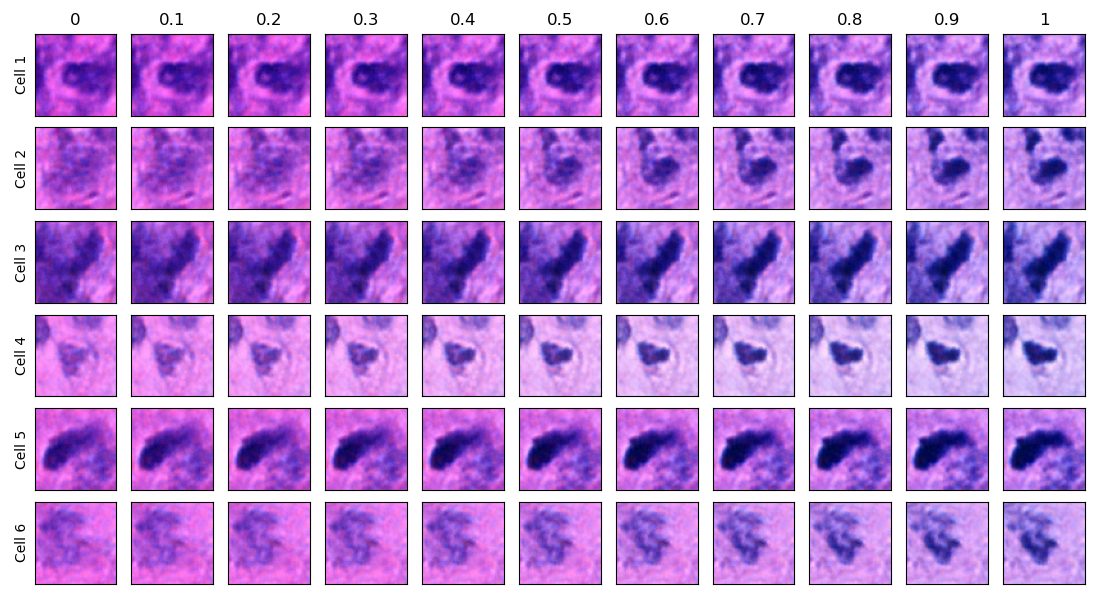}
\caption{Sets of synthetic cells generated with the Meningioma DPM. Each row illustrates a non-mitotic nucleus transitioning to a mitotic figure.} \label{fig4}
\end{figure}
\vspace{-20pt}

\subsection{Meningioma Visual Results}
Fig \ref{fig4} shows selected examples from the DPM trained on the meningioma dataset. 
 In cells 1, 4 and 5, the nucleus size stays roughly the same while the chromatin gets darker and sharper, accentuating the irregularity in the chromatin edges. In cells 2 and 6, the dense chromatin is created from a blurry and indistinct starting image. The sharp, dense and granular chromatin in cell 6 eventually resembles prophase. In cell 3 the intensity of the chromatin increases while elongating and sharpening, resembling metaphase. Note that, the differences between the first and last columns are in general very subtle to the untrained eye.

\subsection{Generating Mitotic Figures from Real Negative Examples}

Mitotic figure variants can also be generated from real negative examples. 
In a DPM, the forward diffusion mode gradually adds noise to an input image over multiple time-points.
Inverting this process, allows us to convert a noisy image into a realistic looking image.
When running the inverse process, we conditioned on a mitosis score of 1, ensuring that the DPM attempted to generate a mitotic figure.
Note that, in our experiments below, we always input a non-mitotic image and run the forward and backward modes for a varying number of time-points. 
The longer we run these modes, the less the output image resembles the input and the more mitotic it looks.

\begin{figure}
\centering
\includegraphics[width=\textwidth]{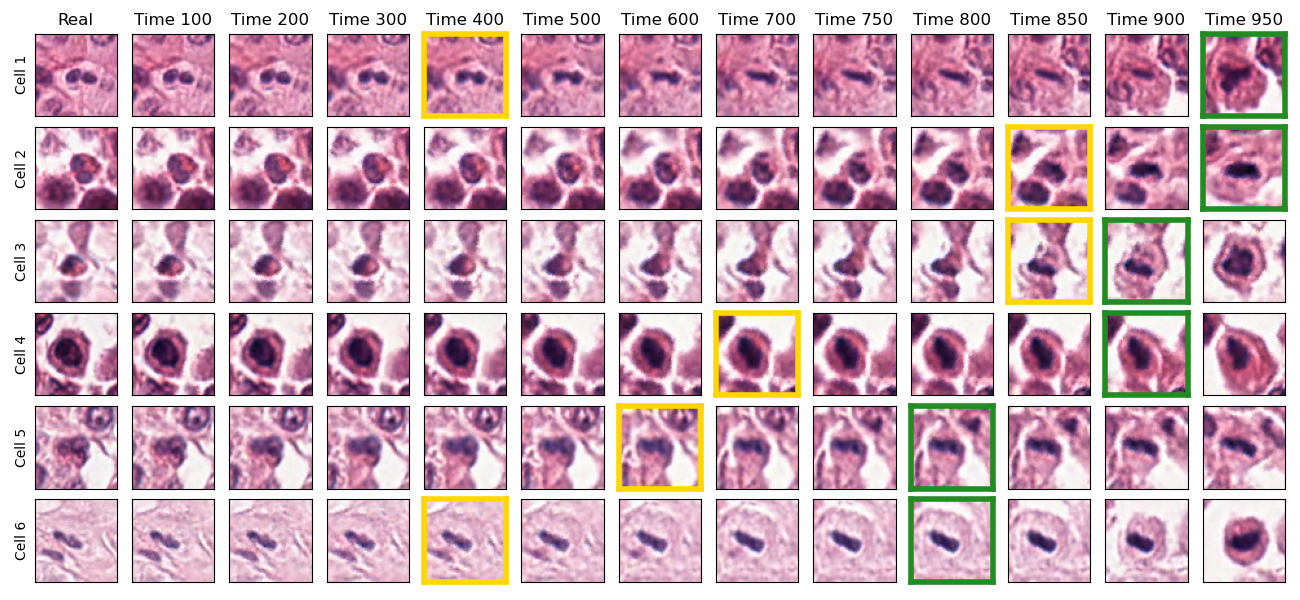}
\caption{Generated Mitotic Figures from real negative examples from CCMCT dataset during intermediate time points.} \label{fig5}
\end{figure}

In Fig \ref{fig5}, 6 negative cell examples from 6 different WSIs are shown in the real column. 
Each column corresponds to a different number of time-points that we used to run in the forward and inverse modes. 
We can appreciate that the real negative cell examples are slowly transitioning into MFs, with a variety of changes occurring in intermediate steps. 
Each cell was annotated by a pathologist retrospectively to mark the earliest time-point where the cell resembles a MF, indicated with the yellow frame, and when the example becomes a convincing MF, indicated with the green frame. 
We can see that the examples that reach the yellow frame earlier, such as cells 1 and 6, already exhibit features associated with mitosis, such as dense and elongated chromatin. 
Conversely, examples that reach the yellow frame later can be thought of as more obvious negative examples. 
Examples such as cell 3, where the jump from the yellow to green state occurs in a single frame, allow us to isolate the changes needed for certainty.
In this case, we observe that the increased definition of the cell body while the chromatin remains the same, is the driving feature. 
Overall, the relative time points in which the real negative cell examples morph into a mitotic figure can also be used as a marker to determine which examples are likely to be mistaken for mitotic figures by classification models, and the data sets can be artificially enhanced for those examples to reduce misclassification.

\vspace{-32pt}

\section{Discussion}

Detecting mitotic figures is a clinically significant, yet difficult and subjective task.
There is no gold-standard ground-truth and we rely on expert pathologists for annotations.
However, the variation and subjectivity of pathologist annotations is well documented. 
To date, there has been little focus on the uncertainty in pathologist labels and no systematic effort to specify the image features that drive a mitosis call.

In this paper, we proposed the use of a diffusion probabilistic model, to characterize the MF detection task.
We presented strategies for visualizing how a (real or synthetic) non-mitotic nucleus can transition into a mitosis state.
We observed that there is a wide variety of features associated with this transition, such as intensity changes, sharpening, increase of irregularity, erasure of certain spots, merging of features, redefinition of the cell membrane and increased granularity of the chromatin or cytoplasm. 

There are several directions we would like to explore in the future. 
The conditional DPM can be used to enrich the training data or generate training datasets comprising only of synthetic images to train more accurate MF detection models, particularly in limited sample size scenarios.
The synthetic images can be used to achieve more useful ground-truth labels that do not force pathologists to an arbitrary consensus, but instead allows them to weigh the different features in order to converge on a probabilistic annotation. 
These probabilistic labels can, in turn, yield more accurate, calibrated and/or robust nucleus classification tools.
We also are keen to extend the DPM to condition on different nucleus types and tissue classes.

\section{Acknowledgements}
Funding for this project was partially provided by The New York-Presbyterian Hospital William Rhodes Center for Glioblastoma—Collaborative Research Initiative,  a Weill Cornell Medicine Neurosurgery-Cornell Biomedical Engineering seed grant, The Burroughs Wellcome Weill Cornell Physician Scientist Program Award, NIH grant R01AG053949, and the NSF CAREER 1748377 grant. Project support for this study was provided by the Center for Translational Pathology of the Department of Pathology and Laboratory Medicine at Weill Cornell Medicine.

%
%

%
%
%
\bibliographystyle{splncs04}
\bibliography{main}

\end{document}